%% file: main.tex
\documentclass[10pt,twocolumn,letterpaper]{article}

\usepackage{unknown}             

\input{preamble}

\definecolor{unknownblue}{rgb}{0.21,0.49,0.74}
\usepackage[pagebackref,breaklinks,colorlinks,citecolor=unknownblue]{hyperref}
\usepackage[accsupp]{axessibility} 

\def\paperID{786} 
\def\confName{unknown}
\def\confYear{unknown}

\title{Real-Time Person Image Synthesis Using a Flow Matching Model}

\author{
Jiwoo Jeong\textsuperscript{\rm 1,2} \and 
Kirok Kim\textsuperscript{\rm 1} \and 
Wooju Kim\textsuperscript{\rm 1 \thanks{Corresponding author.}} \and 
Nam-Joon Kim\textsuperscript{\rm 2 \thanks{Corresponding author.}} \and
\textsuperscript{\rm 1}	Department of Industrial Engineering, Yonsei University\\
\textsuperscript{\rm 2} Next-Generation Semiconductor, Seoul National University\\
{\tt\small \{dailyavenger2,alfmalfm11,wkim\}@yonsei.ac.kr, knj01@snu.ac.kr}
}

\begin{document}
\maketitle
\input{sec/0_abstract}    
\input{sec/1_introduction}
\input{sec/2_related_work}

\input{sec/3_method}

\input{sec/4_experiment}
\input{sec/5_conclusion}

{
    \small
    \bibliographystyle{unknownnat_fullname}
    \bibliography{main}
}


\clearpage
\input{sec/X_suppl}

\end{document}

%% file: preamble.tex
\usepackage[dvipsnames]{xcolor}

\usepackage{makecell}
\usepackage{multirow}
\usepackage{colortbl}
\usepackage{bm}
\usepackage{indentfirst}
\usepackage{float}
\usepackage{algorithm}
\usepackage{algorithmic}
\usepackage{adjustbox}

%% file: sec/0_abstract.tex
\begin{abstract}
Pose-Guided Person Image Synthesis (PGPIS) generates realistic person images conditioned on a target pose and a source image. This task plays a key role in various real-world applications, such as sign language video generation, AR/VR, gaming, and live streaming. In these scenarios, real-time PGPIS is critical for providing immediate visual feedback and maintaining user immersion.However, achieving real-time performance remains a significant challenge due to the complexity of synthesizing high-fidelity images from diverse and dynamic human poses. Recent diffusion-based methods have shown impressive image quality in PGPIS, but their slow sampling speeds hinder deployment in time-sensitive applications. This latency is particularly problematic in tasks like generating sign language videos during live broadcasts, where rapid image updates are required. Moreover, AR/VR, gaming, and live streaming necessitate real-time responsiveness. Therefore, developing a fast and reliable PGPIS model is a crucial step toward enabling real-time interactive systems. To address this challenge, we propose a generative model based on flow matching (FM). Unlike diffusion models, FM employs deterministic updates and avoids the injection of noise in intermediate steps. As a result, our approach enables faster, more stable, and more efficient training and sampling. Furthermore, the proposed model supports conditional generation and can operate in latent space, making it especially suitable for real-time PGPIS applications where both speed and quality are critical. We evaluate our proposed method, Real-Time Person Image Synthesis Using a Flow Matching Model (RPFM), on the widely used DeepFashion dataset for PGPIS tasks. Our results show that RPFM achieves near-real-time sampling speeds while maintaining performance comparable to the state-of-the-art X-MDPT model. Our methodology trades off a slight, acceptable decrease in generated-image accuracy for over a twofold increase in generation speed, thereby ensuring real‑time performance. At 256×256 resolution with a batch size of 8, the fastest model to date—X‑MDPT—requires 1.191s even in its smallest “small” configuration, whereas our medium‑size model completes generation in just 0.489s. In terms of image quality at 512×512 resolution, X‑MDPT achieves an FID of 7.162, an LPIPS of 0.1645, and an SSIM of 0.7522; by contrast, our medium‑size model records an FID of 8.522, an LPIPS of 0.1786, and an SSIM of 0.7742—overall comparable metrics, with an even higher SSIM.
Code is available at \url{https://github.com/SONNY2020-c/RPFM-official-code}.
\end{abstract}

%% file: sec/1_introduction.tex
\section{Introduction}
\label{sec:intro}

The creation of realistic virtual human images constitutes a pivotal area of research within the domains of computer vision and generative models. Among these, Pose-Guided Person Image Synthesis (PGPIS) is particularly focused on generating images that preserve the visual characteristics of a source image while adapting it to a specified target pose~\cite{liao2024appearance}. This technique finds applications in various fields, including virtual reality (VR) and e-commerce~\cite{sha2023deep}, and holds significance for tasks such as person re-identification~\cite{zhu2019progressive} and the generation of sign language videos.

\begin{figure}[t]
  \centering
   \includegraphics[width=0.6\linewidth]{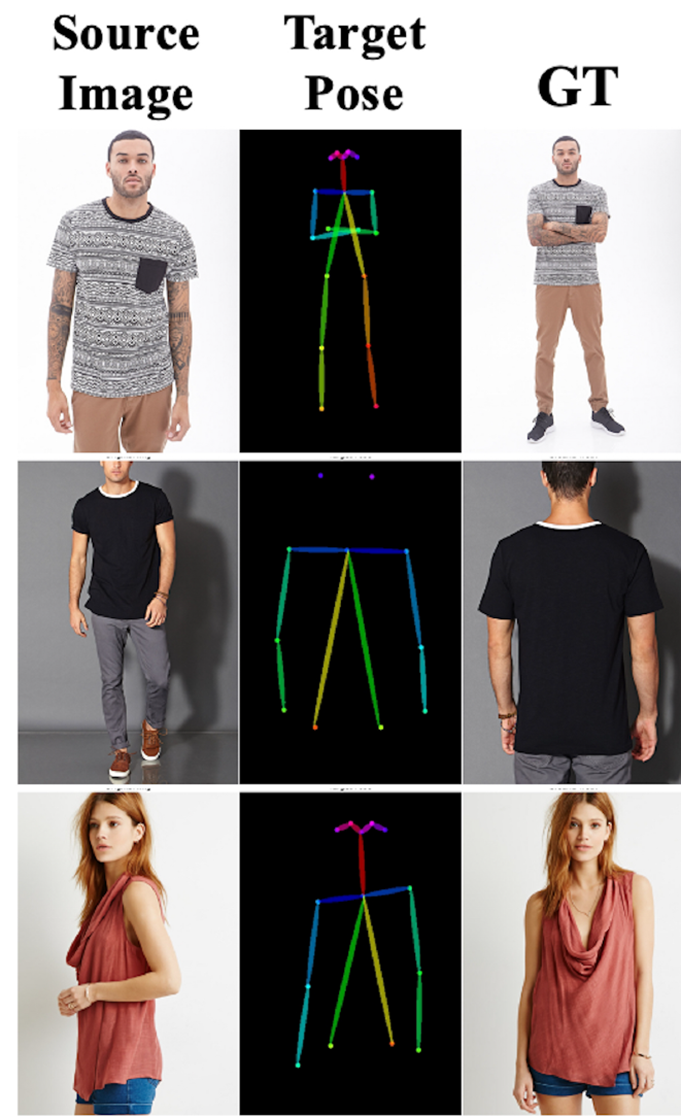}

   \caption{An illustration of the DeepFashion PGPIS dataset is presented. In this task, the individual depicted in the source image must be modified to align with the target pose.}
   \label{fig:deepfashion}
\end{figure}

The PGPIS task entails transforming an individual's image to correspond with a designated pose, necessitating the alignment of the source image with the target pose. This process frequently encounters difficulties in synthesizing occluded regions, where the model must infer parts not visible in the source image. For instance, in the first row of \cref{fig:deepfashion}, the model is required to predict the individual's lower body to generate the target image. 

Recent research in the field of PGPIS has predominantly focused on diffusion-based methodologies, with prominent examples including PIDM~\cite{bhunia2023person}, PCDM~\cite{shen2023advancing}, X-MDPT~\cite{pham2024cross}, and CFLD~\cite{lu2024coarse}. Nevertheless, these models encounter challenges related to slow sampling speeds.

Diffusion~\cite{ho2020denoising} is a score-based model that employs Stochastic Differential Equations (SDEs), necessitating hundreds to thousands of function evaluations for sampling. This requirement significantly decelerates the process, rendering it computationally inefficient for practical applications.

Continuous Normalizing Flows (CNFs)~\cite{chen2018neural} have recently emerged as a significant category within generative models, capable of modeling arbitrary probability paths. The Flow Matching model facilitates the training of CNFs without the need for simulation~\cite{lipman2022flow}. Furthermore, by mapping images to a latent space, generative models can be effectively applied within this space, thereby enabling conditional generation through flow matching~\cite{dao2023flow}.

Flow matching seeks to learn an ordinary differential equation that delineates a trajectory from a source distribution to a target distribution. Generative models based on ordinary differential equations, such as this, are recognized for offering more rapid training and sampling speeds compared to stochastic differential equation-based models, such as diffusion models.~\cite{dao2023flow}

While flow matching has been recently investigated for real-time generation within the speech domain~\cite{le2024voicebox, liu2023generative}, its application in this specific area remains unexplored. The Motion Flow Matching~\cite{hu2023motion} paper generates poses from textual input; however, it predominantly produces mannequin-like figures, failing to preserve visual appearance or maintain a specified pose condition. Consequently, this represents a distinct task from PGPIS. We propose the application of flow matching to develop a conditional generation model for PGPIS, with the objective of enhancing sampling speed, and we substantiate our approach through experimental validation.

Our major contributions are as follows:
\begin{itemize}
\item We propose a transformer-based flow matching model for pose-guided person image synthesis, where a pre-trained VAE maps images to a latent space and a DiT-based neural network predicts the latent flow. By conditioning on both source images and target poses, our method efficiently aligns the source to the desired pose. This design enables significant reduction in inference steps, achieving near real-time image and video generation without compromising output quality.
\item  We introduce two complementary conditioning techniques—input concatenation and conditional aggregation—to enhance both the speed and performance of conditional generation in our model. Input concatenation merges the noisy latent, source image, and target pose into a unified representation, enabling the model to directly incorporate spatial features with minimal computational overhead. Conditional aggregation integrates four distinct condition embeddings (local/global features of both source image and target pose) through transformer-based self- and cross-attention. This architecture allows for efficient conditioning, thereby reducing inference time. Our design extends previous works that used only three conditional embeddings, and ablation studies demonstrate that the combined use of both methods not only improves generation quality but also accelerates the PGPIS model, making it suitable for real-time applications.
\end{itemize}

%% file: sec/2_related_work.tex
\section{Related Work}
\label{sec:related work}

\subsection{Neural ODE}

Neural Ordinary Differential Equations (Neural ODEs)~\cite{chen2018neural} constitute an innovative class of deep neural networks. Conventional deep learning architectures, such as ResNet, consist of multiple discrete, independent layers, with each layer processing values at a single time step. Neural ODEs draw inspiration from the resemblance between the residual connections in ResNet and Euler’s method in solving ordinary differential equations, both of which involve the accumulation of incremental changes. Unlike ResNet~\cite{he2016deep} models, which are limited by a finite number of layers (e.g., ResNet50 or ResNet152), Neural ODEs theoretically permit an infinite number of layers by utilizing Euler’s method with smaller step sizes, thereby facilitating a transition from discrete to continuous transformations.

Neural Ordinary Differential Equations (ODEs) employ a continuous-depth residual network alongside a continuous-time latent variable model. In contrast to traditional ResNet, which utilizes equations such as \cref{eq:resnet},
\begin{equation}
  h_{t+1} = h_t + f(h_t, \theta_t)
  \label{eq:resnet}
\end{equation}
Neural ODEs make use of equations like \cref{eq:NODE}. 
\begin{equation}
  \frac{dh(t)}{dt} = f(h(t), t, \theta) %
  \label{eq:NODE}
\end{equation}
Additionally, Neural ODEs are applied in continuous normalizing flows. They can be implemented using ODE solvers and facilitate backpropagation through the adjoint sensitivity method.

\subsection{Flow Matching Model}

The fundamental concept underlying flow-based generative models~\cite{rezende2015variational} is the optimal delineation of a trajectory that transforms a simple distribution, such as a standard Gaussian, into the actual data distribution, while ensuring the reversibility of this transformation. Prior to the development of flow matching methodologies, generative models employed discrete normalizing flows or continuous normalizing flows via Neural ODEs~\cite{chen2018neural} to simulate changes in probability distributions. However, these approaches necessitated the simulation of the data distribution through a neural network, which required continuous optimization by comparing it to the true data distribution, thereby incurring significant computational costs.

Utilizing the flow matching approach~\cite{lipman2022flow}, it is posited that if a neural network is capable of capturing the flow of a data distribution $dx/dt$, the alterations in the probability distribution can be ascertained through the application of the continuity equation, as derived from fluid dynamics.

In diffusion models~\cite{ho2020denoising}, each timestep in the forward diffusion process incorporates a stochastic element, introducing inherent randomness at each stage. Conversely, in flow matching, transitions are executed deterministically at each timestep. Once the initial Gaussian noise is established, the trajectory toward the generated image becomes deterministic.

This method facilitates flow matching to produce a more efficient sampling trajectory compared to diffusion models, thereby decreasing the number of inference steps necessary during sampling. Furthermore, as demonstrated in the ~\cite{dao2023flow}, flow matching can be executed within the latent space for image generation, enhancing its effectiveness. Conditional generation is also feasible, contingent upon the architectural framework.

\subsection{Pose Guided Person Image Synthesis}

Since the introduction of the Pose-Guided Person Image Synthesis (PGPIS) model in the ~\cite{ma2017pose}, significant research has been conducted in this domain. Recently, diffusion-based PGPIS studies have gained prominence, achieving state-of-the-art results. The pioneering work utilizing diffusion was PIDM~\cite{bhunia2023person}, which employed a latent diffusion model (LDM). Subsequently, various methodologies have been developed to enhance performance: CFLD~\cite{lu2024coarse}, which incorporates an attention module for feature extraction; X-MDPT~\cite{pham2024cross}, which utilizes a transformer model; and PCDM~\cite{shen2023advancing}, which predicts the embedding of the target image as a condition and includes a refining stage. However, due to their reliance on diffusion, these models exhibit low inference speed, thereby limiting their efficacy for real-time image and video generation. To address this limitation, this paper proposes conditional image generation based on flow matching.

%% file: sec/3_method.tex
\section{Method}
\label{sec:method}

\subsection{Preliminary}

To delineate the transition from a standard Gaussian distribution to the empirical data distribution, the flow matching model~\cite{lipman2022flow} employs a neural network to predict the flow $dx/dt$ of the data distribution. Initially, the timestep is defined within the interval [0, 1]. During the training phase, $x_1$ at $t=1$ is sampled from the empirical dataset, while $x_0$ at $t=0$ is sampled from the initial Gaussian distribution. Following the computation of $dx/dt$, the neural network is trained to predict this flow. In the inference phase, $x_0$ is sampled from the initial Gaussian distribution, and the trained model provides $dx/dt$ for t within the range of 0 to 1. Subsequently, an ODE solver computes $x_1$ to facilitate the sampling of the generated data.

Refer to \cref{main_algorithm} and \cref{main_algorithm2} for the operational details of the algorithm. In Algorithm 1, at line 4, the timestep is uniformly sampled. Subsequently, at line 5, $x_1$ is sampled from the dataset. At line 6, $x_0$ is sampled from noise, and at line 7, $x_t$ is generated through the conditional flow.

In Algorithm 2, at line 2, the initial value $x_0$ is sampled from noise, and the subsequent image $x_1$ is generated using an ODE solver.

\begin{figure}[h!]
\vspace{-0.5em}
\centering
\begin{algorithm}[H]
\caption{Flow Matching training}
\label{main_algorithm}
\begin{algorithmic}[1]
\STATE \textbf{Input:} dataset $q$, noise $p$
\STATE Initialize $v^{\theta}$
\WHILE{not converged \textbf{do}}
    \STATE $t \sim \mathcal{U}([0,1])$
    \STATE $x_1 \sim q(x_1)$
    \STATE $x_0 \sim p(x_0)$
    \STATE $x_t=\mathcal{\psi}_t(x_0|x_1)$
    \STATE Gradient step with $\nabla_{\theta}\Big\| v_{t}^{\theta}(x_t)-(\dot{x_t}) \Big\|^2$
\ENDWHILE
\STATE \textbf{Output:} $v^{\theta}$    
\end{algorithmic}
\end{algorithm}
\vspace{-1.5em}
\end{figure}

\begin{figure}[h!]
\vspace{-0.5em}
\centering
\begin{algorithm}[H]
\caption{Flow Matching sampling}
\label{main_algorithm2}
\begin{algorithmic}[1]
\STATE \textbf{Input:} trained model $v^{\theta}$    
\STATE $x_0 \sim p(x_0)$
\STATE Numerically solve ODE $(\dot{x_t})=v_{t}^{\theta}(x_t)$
\STATE \textbf{Output:} $x_1$
\end{algorithmic}
\end{algorithm}
\vspace{-1.5em}
\end{figure}

In the context of flow matching~\cite{lipman2022flow}, the conditional optimal transport path (Cond-OT) is conventionally employed to determine the trajectory $dx/dt$. In essence, this trajectory is defined in a linear and optimal manner using the expression $x_t=(1-t)x_0+tx_1$, which facilitates rapid and straightforward sampling. In practical applications, the following equations are utilized.
\begin{equation}
  \psi_t(x) = (1-(1-\sigma_{\text{min}})t)x + tx_1
  \label{eq:FM1}
\end{equation}
{
\begin{equation}
  \mathcal{L}_{\scriptscriptstyle \text{FM}}(\theta) 
     =\mathbb{E}_{t, q(x_1), p(x_0)} \Big\|v_t(\psi_t(x_0)) - \Big (x_1 - (1-\sigma_{\min})x_0\Big) \Big\|^2
  \label{eq:FM2}
\end{equation}
}
where $\sigma_{\min}$ is almost zero.

In the study of Flow Matching in Latent Space (LFM)~\cite{dao2023flow}, akin to the Stable Diffusion~\cite{rombach2022high} approach, images are transformed into a lower-dimensional space via a Variational Autoencoder (VAE)~\cite{esser2021taming}. The generative model is subsequently trained and inferred within this latent space before being reconverted to the image space to produce the final image. Depending on the architectural configuration, the model may also accommodate various conditions.

\subsection{Flow Matching Model}

\begin{figure*}
  \centering
  \includegraphics[width=1\linewidth]{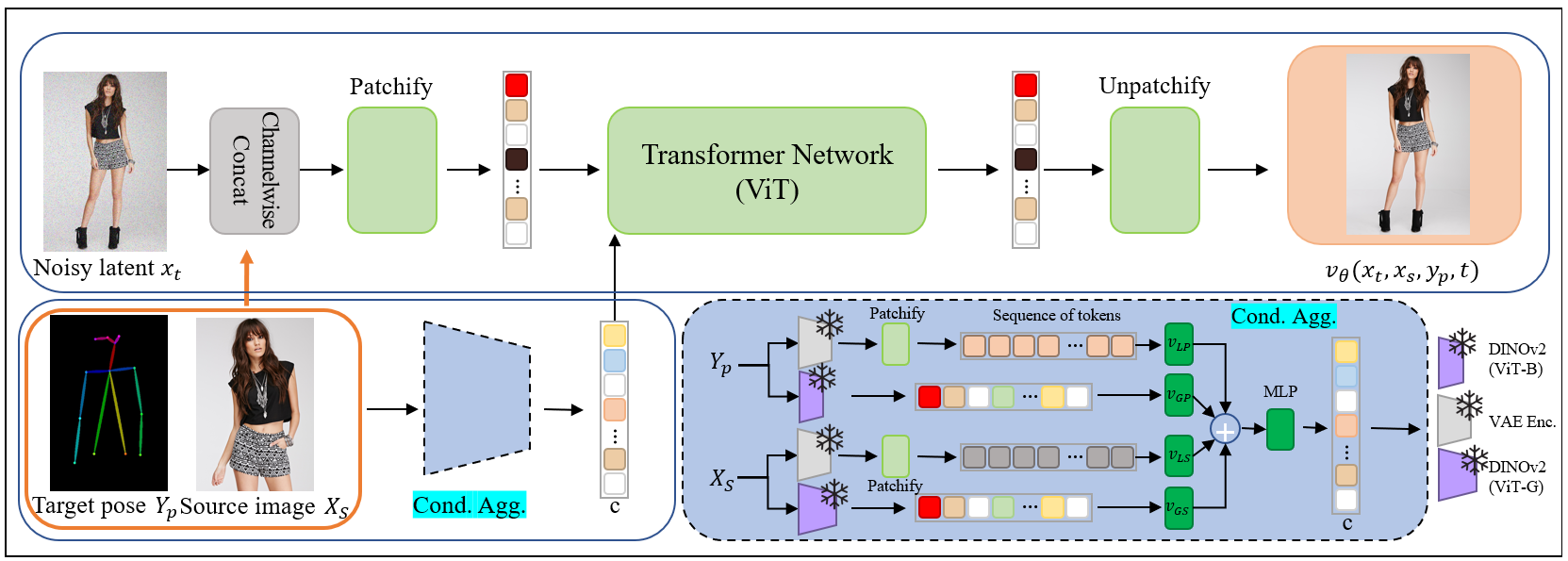}
  \caption{The architecture of our Real-Time Person Image Synthesis employs a Flow Matching Model. This architecture encompasses both the input concatenation component and the conditional aggregation component, with the output represented as $dz/dt$. It is important to note that certain VAE components have been excluded from the architectural diagram, specifically the inputs and conditions intended for concatenation at the input concatenation stage. The orange arrow indicates input concatenation.}
  \label{fig:architecture}
\end{figure*}

We introduce RPFM (Real-time Person Image Synthesis using a Flow Matching Model). The architecture of our model is depicted in \cref{fig:architecture}.

\textbf{Motivation.} Prior research on pose-guided person synthesis has predominantly utilized methods based on Generative Adversarial Networks (GANs)~\cite{goodfellow2020generative} or diffusion processes. GAN-based models~\cite{ren2022neural, zhang2022exploring, zhou2022cross} endeavor to transform the style of a source image to align with a target pose in a single forward pass, which frequently results in suboptimal performance and unstable training dynamics~\cite{iglesias2023survey}. Conversely, diffusion-based models~\cite{bhunia2023person, lu2024coarse} attain more stable training and enhanced performance through multiple conditional denoising diffusion steps. Nonetheless, a notable limitation of diffusion-based models is their slow inference speed, attributed to the numerous steps involved. For instance, in the context of generating sign language videos, where a pose sequence is converted into an image sequence on a frame-by-frame basis, the slow sampling speed can pose a significant challenge.

In response to this issue, we have developed a flow matching model, which can be conceptualized as a generalized framework that includes diffusion models. The flow matching model enhances inference speed by decreasing the average number of function evaluations (NFE), specifically by minimizing the frequency of neural network calls during the sampling process. This methodology is particularly advantageous in contexts requiring real-time data generation and has already been implemented in the speech domain~\cite{le2024voicebox, liu2023generative}.

We developed a model that effectively reduces NFE while preserving performance, thereby achieving a sampling speed conducive to near real-time video generation.

Drawing upon the findings of the LFM~\cite{dao2023flow} paper, which demonstrated that employing DiT~\cite{peebles2023scalable} as the backbone model results in reduced inference time compared to UNet~\cite{ronneberger2015u}, we have developed a transformer-based flow matching model. In this model, the source image, target pose image, and noisy latent are concatenated as inputs (input concatenation), while various embeddings of the source image and target pose image are aggregated as conditions for DiT (conditional aggregation).

\textbf{Transformer-based Flow Matching Model.}
In our RPFM model, we have structured the neural network utilizing the DiT model as a foundation. Conventionally, diffusion models employ convolutional U-Nets as the backbone for image generation. Nevertheless, the DiT paper illustrates that employing a transformer as the backbone model can also yield effective results.

Consider a scenario where we are provided with a source image $X_s\in\mathbb{R}^{256\times 256\times 3}$ and a target pose image $Y_p\in\mathbb{R}^{256\times 256\times 3}$. Our objective is to train a backbone model $\theta$ to align the source image with the target pose, thereby generating the target image $Y$. Initially, a pre-trained Variational Autoencoder (VAE) is employed to transform pixel images into latent representations, yielding $x_s\in\mathbb{R}^{32\times 32\times 4}$, $y_p\in\mathbb{R}^{32\times 32\times 4}$, and $y\in\mathbb{R}^{32\times 32\times 4}$. Subsequently, a neural network, which is based on a transformer architecture, predicts $dz/dt$ (where $z$ represents the noisy latent variable, with $z_0$ originating from $\mathcal{N}(0,\mathbf{I})$ and $z_1$ being $y$). This procedure is designed to map the initial Gaussian distribution $\mathcal{N}(0,\mathbf{I})$ to the target image distribution, utilizing $X_s$ and $Y_p$ as conditioning inputs. The loss function employed is based on Mean Squared Error (MSE) Loss and is computed as follows.
\begin{equation}
  \psi_t(z_0) = (1-(1-\sigma_{\text{min}})t)z_0 + ty
  \label{eq:RPFM1_1}
\end{equation}
\begin{equation}
  \begin{split}
    \mathcal{L}_{\scriptscriptstyle \text{FM}}(\theta) 
       =\mathbb{E}_{t, y, p(z_0)\sim\mathcal{N}(0,\mathbf{I}), x_s, y_p} \\ 
         &\Big\|v_\theta(\psi_t(z_0),x_s,y_p,t) \\
         & - \Big (y - (1-\sigma_{\min})z_0\Big) \Big\|^2.
    \label{eq:RPFM1_2}
  \end{split}
\end{equation}
Following the training of the neural network, the process of sampling the target image commences with the extraction of noisy data from the initial Gaussian distribution $\mathcal{N}(0,\mathbf{I})$. By employing $x_s$ and $y_p$ as conditions, we derive $dx/dt$, which is subsequently resolved using an ODE solver to generate the target image $Y$. The design of input concatenation and conditional aggregation aims to enhance the performance of flow matching, aligning it more closely with that of diffusion. In the DiT architecture, conditions are integrated with the input through self-attention and cross-attention mechanisms, thereby enabling multiple layers of the transformer model to effectively reference the condition.

\textbf{Input Concatenation.} This section integrates three distinct inputs. The initial input, denoted as the noisy latent $z_t$, represents the noisy latent state at timestep t that directs the model towards the desired image. The second input, $x_s$, is derived by processing the source image through the VAE encoder. The third input, $y_p$, is obtained by encoding the target pose image via the VAE encoder. All three inputs share identical dimensions $32 \times 32 \times 4$, resulting in a combined input of dimensions $32 \times 32 \times 12$ following channel-wise concatenation. This concatenated input is subsequently processed through the patchify layer and introduced into the Transformer Network. This configuration enables the model to integrate the conditions alongside the noisy latent to generate the output. For the pose image, a 3-channel RGB pose image of size $256 \times 256 \times 3$ is utilized. The orange arrow in \cref{fig:architecture} illustrates the concatenation of conditions into the input. In the ablation study, it was demonstrated that input concatenation enhances performance.

\textbf{Conditional Aggregation.} This section undertakes the aggregation of four distinct conditions. The first condition, $v_{\text{LS}}$ (Local Source Image Embedding), represents the patchified embedding of the source image feature, as obtained from the VAE encoder. The second condition, $v_{\text{GS}}$ (Global Source Image Embedding), is derived from the source image utilizing a pretrained DINOv2 encoder~\cite{oquab2023dinov2}. The third condition, $v_{\text{LP}}$ (Local Target Pose Image Embedding), pertains to the target pose image feature acquired from the VAE encoder. The fourth condition, $v_{\text{GP}}$ (Global Target Pose Image Embedding), is derived from the target pose image using a pretrained DINOv2 encoder. In \cref{fig:architecture}, the black arrow entering the Transformer Network signifies the outcome of conditional aggregation, which is employed as the condition.

Local Feature: The latent image, characterized by dimensions of $32 \times 32 \times 4$, undergoes processing via the patchify layer, resulting in the generation of 256 tokens, each with embeddings of size $256 \times D$, where D denotes the embedding dimension. Subsequently, a $1 \times 1$ convolution layer is employed to reduce the 256 channels to a single channel, thereby forming a local vector with a dimension of D.

Global Feature: The DINOv2-B encoder is employed for the target pose image, while the DINOv2-G encoder is utilized for the source image condition. Both images, originally sized at $256 \times 256 \times 3$, are resized to $224 \times 224 \times 3$ before being processed through the DINOv2 encoder. The encoder generates a sequence of embeddings comprising 257 tokens, inclusive of the CLS token, with dimensions of $257 \times D$. Subsequently, a $1 \times 1$ convolution layer reduces the 257 channels to a single channel, resulting in a global vector of dimension D.

Aggregation: We aggregate the four output vectors through an addition operation to derive a D-dimensional vector, which is subsequently processed through the final MLP layer to yield the ultimate condition vector of dimension D. Our ablation study demonstrated that conditional aggregation significantly enhances performance.

\textbf{Classifier-Free Guidance.} In the process of image generation, we utilized classifier-free guidance~\cite{ho2022classifier}, a technique frequently employed in both diffusion and flow matching models. The inference procedure is executed as demonstrated in the following equation:
\begin{equation}
  \hat{v}_\theta(z_t,c,t) = v_\theta(z_t,t) + \gamma_t (v_\theta(z_t,c,t) - v_\theta(z_t,t)).
  \label{eq:RPFM_CFG}
\end{equation}
where $v_\theta(z_t,t)$ refers to the scenario involving input concatenation without the application of conditional aggregation, whereas $v_\theta(z_t,c,t)$ pertains to the scenario incorporating both input concatenation and conditional aggregation.

%% file: sec/4_experiment.tex
\section{Experiment Result}
\label{sec:experiment}

\subsection{Implementation}

\textbf{Datasets.} We conducted our experiments utilizing images with a resolution of 256x256 from the DeepFashion In-shop Clothes Retrieval Benchmark dataset~\cite{liu2016deepfashion}, and we compared the performance of our method against baseline studies. The DeepFashion dataset comprises 52,712 high-resolution images of individuals, specifically curated for the fashion domain. This dataset is divided into non-overlapping training and testing sets, consisting of 101,966 pairs and 8,570 pairs, respectively. We extracted skeletons using OpenPose and visualized pose images as RGB images employing 18 key points.

\textbf{Metrics.} We assessed the images produced by our model utilizing four distinct metrics: FID~\cite{heusel2017gans}, LPIPS~\cite{zhang2018unreasonable}, SSIM~\cite{wang2004image}, and PSNR. Both FID and LPIPS are predicated on deep features. FID calculates the Wasserstein-2 distance between the distributions of generated and real images using Inception-v3. LPIPS employs a network trained on human judgment to evaluate reconstruction accuracy. SSIM and PSNR measure the similarity between generated images and ground truth at the pixel level.

\textbf{Implementation Details.} We employed PyTorch in conjunction with pre-trained VAE and DINO-v2 models, conducting our experiments on eight A100 or H100 GPUs (80G). For medium-sized models, the batch size was set to 240, while for large-sized models, it was set to 104. The models were trained for 500 epochs with a fixed learning rate of 1e-4, utilizing the AdamW optimizer. The original images in the DeepFashion dataset, which have a resolution of 256x176, were resized to 256x256 for training and sampling with our generative model. In the ablation study, due to resource constraints, each experiment was limited to 100 epochs to facilitate performance comparison and the selection of the most efficient model. For classifier-free guidance, the optimal scale value was determined to be 1.1, and we will subsequently demonstrate the performance variation based on the cfg scale value later in the paper.

\subsection{Main Results}

\begin{table}[h]
\centering
\renewcommand\arraystretch{1.2}
\begin{adjustbox}{width=0.5\textwidth,center}
\begin{tabular}{c|cccc} 
\toprule
Method & FID($\downarrow$) & LPIPS($\downarrow$) & SSIM($\uparrow$) & PSNR($\uparrow$) \\ 
\midrule
SPGNet~\cite{lv2021learning} & 16.184 & 0.2256 & 0.6965 & 17.222 \\
DPTN~\cite{zhang2022exploring} & 17.419 & 0.2093 & 0.6975 & 17.811 \\
NTED~\cite{ren2022neural} & 8.517 & 0.1770 & 0.7156 & 17.740 \\
CASD~\cite{zhou2022cross} & 13.137 & 0.1781 & 0.7224 & \underline{17.880} \\
PIDM~\cite{bhunia2023person} & \underline{6.812} & 0.2006 & 0.6621 & 15.630 \\
PoCoLD~\cite{han2023controllable} & 8.067 & 0.1642 & 0.7310 & - \\
CFLD~\cite{lu2024coarse} & \textbf{6.804} & \textbf{0.1519} & \textbf{0.7378} & \textbf{18.235} \\
X-MDPT-L~\cite{pham2024cross} & 7.287 & 0.1589 & \textbf{0.7405} & - \\
\hline
Ours[Large](NFE=60) & 9.758 & \underline{0.1561} & 0.7360 & 17.825 \\
Ours[Medium](NFE=65) & 8.425 & 0.1591 & 0.7296 & 17.817 \\
\bottomrule
\end{tabular}
\end{adjustbox}
\caption{We performed a quantitative analysis using the DeepFashion dataset, comparing our model with state-of-the-art models such as SPGNet, DPTN, NTED, CASD, PIDM, PoCoLD, CFLD, and X-MDPT. The results for all models, except our own, were sourced from the CFLD or X-MDPT publications. All experiments were conducted on images with a resolution of 256x256. Following this, performance was assessed at a resolution of 256x176.}
\label{tab:result}
\vspace{-1.0em}
\end{table}

\begin{table}[h]
\centering
\renewcommand\arraystretch{1.2}
\begin{adjustbox}{width=0.5\textwidth,center}
\begin{tabular}{c|cccc} 
\toprule
Method & FID($\downarrow$) & LPIPS($\downarrow$) & SSIM($\uparrow$) & PSNR($\uparrow$) \\ 
\midrule
CoCosNet2~\cite{zhou2021cocosnet} & 13.325 & 0.2265 & 0.7236 & - \\
NTED~\cite{ren2022neural} & 7.645 & 0.1999 & 0.7359 & 17.385 \\
PoCoLD~\cite{han2023controllable} & 8.416 & 0.1920 & 0.7430 & - \\
CFLD~\cite{lu2024coarse} & \textbf{7.149} & 0.1819 & 0.7478 & \underline{17.645} \\
X-MDPT-L~\cite{pham2024cross} & \underline{7.162} & \textbf{0.1645} & \underline{0.7522} & - \\
\hline
Ours[Medium](NFE=55) & 8.522 & \underline{0.1786} & \textbf{0.7742} & \textbf{18.093} \\
\bottomrule
\end{tabular}
\end{adjustbox}
\caption{We performed a quantitative analysis using the DeepFashion dataset, comparing our model with state-of-the-art models such as CoCosNet2, NTED, PoCoLD, CFLD, and X-MDPT. The results for all models, except our own, were sourced from the CFLD or X-MDPT publications. All experiments were conducted on images with a resolution of 512x512. Following sampling at 512x352, the performance was assessed at 256x176.}
\label{tab:result222}
\vspace{-1.0em}
\end{table}

We conducted a quantitative evaluation of our method in comparison with existing approaches, including SPGNet, DPTN, CoCosNet2, NTED, CASD, PIDM, PoCoLD, CFLD, and X-MDPT, which served as baselines. The performance metrics for these models, excluding our method, were sourced from the CFLD or X-MDPT paper. As illustrated in \cref{tab:result} and \cref{tab:result222}, our model demonstrates performance comparable to state-of-the-art models, despite its enhanced sampling speed. For experiments conducted at a resolution of 256×256, the cfg scale was set to 1.1, whereas for those at a resolution of 512×512, it was set to 1.0. In the 256×256 resolution experiments, although the FID score is marginally lower than those of PIDM and CFLD, our large model achieves the second-lowest LPIPS score and the third-highest SSIM score. Furthermore, our large model ranks third in terms of PSNR. The medium model's performance is nearly equivalent to that of the large model, even slightly surpassing it in terms of FID. In the 512×512 resolution experiments, while the FID did not attain state-of-the-art levels, it achieved the second-best LPIPS score, as well as the highest SSIM and PSNR scores.

\subsection{Sampling Speed}

\begin{table}[h]
\centering
\renewcommand\arraystretch{1.2}
\begin{adjustbox}{width=0.4\textwidth,center}
\begin{tabular}{c|c} 
\toprule
Method & Inference Time \\ 
\midrule
PIDM~\cite{bhunia2023person} & 16.975$\pm$0.055s \\
CFLD~\cite{lu2024coarse} & 12.132$\pm$0.064s \\
X-MDPT-S~\cite{pham2024cross} & 1.191$\pm$0.021s \\
X-MDPT-B~\cite{pham2024cross} & 1.299$\pm$0.022s \\
X-MDPT-L~\cite{pham2024cross} & 3.124$\pm$0.026s \\
\hline
Ours[Medium](NFE=60) & 0.489$\pm$0.018s \\
Ours[Large](NFE=30) & 0.550$\pm$0.021s \\ 
Ours[Large](NFE=60) & 0.850$\pm$0.022s \\
\bottomrule
\end{tabular}
\end{adjustbox}
\caption{Comparison of sampling speed was conducted using images with a resolution of 256x256 pixels. Models based on U-Net architectures, such as PoCoLD, were excluded from this comparison due to their inherently slower performance relative to DiT-based models.}
\label{tab:speed}
\vspace{-1.0em}
\end{table}

\begin{figure}[t]
  \centering
   \includegraphics[width=1.2\linewidth]{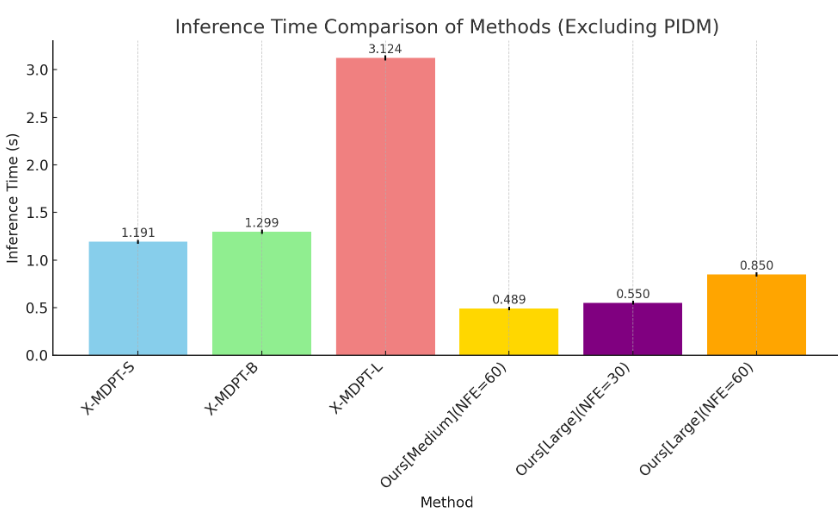}

   \caption{visualization of ~\cref{tab:speed}}
   \label{fig:graph}
\end{figure}

In \cref{tab:speed}, we directly measured the inference time for our model, while the inference times for PIDM and X-MDPT were sourced from the X-MDPT paper. Additionally, we measured the speed of CFLD. Adhering to the measurement protocol outlined in the X-MDPT paper, we assessed the time required to generate eight images concurrently on a single A100 GPU with a batch size of eight, repeating this process ten times to calculate the average. NFE denotes the averaged number of function evaluations. The results indicate that our model exhibits superior speed. The implementation of a flow matching model is a significant factor contributing to this disparity in speed. Furthermore, our experiments revealed that as the batch size increases, the average generation time per image (sampling time divided by batch size) decreases for our model. This suggests that our model is capable of real-time video generation on a frame-by-frame basis. The experimental results can also be verified through the graph below (see \cref{fig:graph}).

\subsection{Ablation Study}

To substantiate the assertion that the concurrent utilization of input concatenation and conditional aggregation structures yields optimal performance, we conducted training on three distinct architectures using a medium-sized model over 100 epochs each. All experiments were executed with the NFE set to 60. The results, as presented in \cref{tab:ab3}, were sampled with a cfg scale of 1.1 across all scenarios. The term "Ours" denotes the architecture delineated in \cref{fig:architecture}. "Ablation1" represents a variant devoid of the input concatenation component from \cref{fig:architecture}, wherein only the noisy latent is introduced into the transformer network, and the embeddings of the target pose and source image are aggregated as conditions, effectively removing the orange arrow in \cref{fig:architecture}. "Ablation2" omits the conditional aggregation component from \cref{fig:architecture}. In this configuration, the noisy latent, target pose, and source image are concatenated as input to the transformer network, with only the timestep employed as a condition. In "Ablation2," the cfg scale is ineffectual, as elucidated in the classifier-free guidance section of the methodology. Despite each architecture being trained for only 100 epochs—a relatively brief duration—the comparative analysis indicates that "Ours" achieved superior training outcomes, thereby affirming the indispensability of both the input concatenation and conditional aggregation components in our model.

We conducted an evaluation of the performance differential between utilizing solely the three condition branches from the X-MDPT~\cite{pham2024cross} paper and employing all four condition branches within the conditional aggregation component of our architecture. To assess their capacity for generating fine-grained images, all experiments were executed with the NFE set to 720. Both architectures were trained using a medium-sized model for 100 epochs each. The results, as presented in \cref{tab:ab4}, were sampled with a cfg scale of 1.1 across all scenarios. "Ours" refers to the architecture delineated in \cref{fig:architecture}. The "three conditional branch" configuration excludes the $V_LP$ component from the conditional aggregation part of our architecture, as utilized in X-MDPT~\cite{pham2024cross}. Despite each architecture being trained for only 100 epochs—a relatively brief duration—the comparison indicates that "Ours" achieved superior training outcomes, thereby substantiating the necessity of four conditional branches in the conditional aggregation component. Our model demonstrates slightly lower SSIM and PSNR values—by approximately 0.0006 and 0.038, respectively—yet achieves superior LPIPS and, most notably, exhibits a marked FID improvement of 0.452.

We conducted an analysis of the impact of the cfg scale on model performance. \cref{tab:ab1} presents the results for the large model trained over 500 epochs with varying cfg scales: 1.0, 1.1, 1.25, and 1.4. Although the SSIM is marginally higher at a cfg scale of 1.25 and the PSNR is higher at a cfg scale of 1.0 compared to 1.1, the cfg scale of 1.1 emerged as the optimal choice across multiple evaluation metrics.

\cref{tab:ab2} presents the outcomes for the medium-sized model trained over 500 epochs, evaluated with configuration (cfg) scales of 1.0, 1.1, and 1.25. While the Structural Similarity Index Measure (SSIM) is higher at a cfg scale of 1.25 and the Peak Signal-to-Noise Ratio (PSNR) is higher at a cfg scale of 1.0 compared to a cfg scale of 1.1, the cfg scale of 1.1 was generally determined to be the optimal choice, aligning with the findings from the large model.

\begin{table}[h]
\centering
\renewcommand\arraystretch{1.2}
\begin{adjustbox}{width=0.5\textwidth,center}
\begin{tabular}{c|cccc} 
\toprule
CFG scale & FID($\downarrow$) & LPIPS($\downarrow$) & SSIM($\uparrow$) & PSNR($\uparrow$) \\ 
\midrule
1.0 & 10.663 & 0.1564 & 0.7327 & 17.973 \\
1.1 & 9.758 & 0.1561 & 0.7360 & 17.825 \\
1.25 & 9.903 & 0.1585 & 0.7363 & 17.530 \\
1.4 & 11.869 & 0.1638 & 0.7320 & 17.222 \\
\bottomrule
\end{tabular}
\end{adjustbox}
\caption{Performance evaluation and analysis were conducted with varying CFG scales. The experiments utilized a large-scale model and employed images with a resolution of 256x256.}
\label{tab:ab1}
\vspace{-1.0em}
\end{table}

\begin{table}[h]
\centering
\renewcommand\arraystretch{1.2}
\begin{adjustbox}{width=0.5\textwidth,center}
\begin{tabular}{c|cccc} 
\toprule
CFG scale & FID($\downarrow$) & LPIPS($\downarrow$) & SSIM($\uparrow$) & PSNR($\uparrow$) \\ 
\midrule
1.0 & 9.181 & 0.1597 & 0.7260 & 17.897 \\
1.1 & 8.425 & 0.1591 & 0.7296 & 17.817 \\
1.25 & 8.696 & 0.1606 & 0.7313 & 17.628 \\
\bottomrule
\end{tabular}
\end{adjustbox}
\caption{A comparative performance analysis was conducted with varying CFG scales. The experiments utilized a medium-sized model and employed images with a resolution of 256x256.}
\label{tab:ab2}
\vspace{-1.0em}
\end{table}

\begin{table}[h]
\centering
\renewcommand\arraystretch{1.2}
\begin{adjustbox}{width=0.4\textwidth,center}
\begin{tabular}{c|ccc} 
\toprule
Method & LPIPS($\downarrow$) & SSIM($\uparrow$) & PSNR($\uparrow$) \\ 
\midrule
Ablation1 & 0.2274 & 0.6651 & 15.300 \\
Ablation2 & 0.2478 & 0.6720 & 15.354 \\
Ours & 0.2102 & 0.6858 & 16.070 \\
\bottomrule
\end{tabular}
\end{adjustbox}
\caption{The study presents a comparative performance analysis of input concatenation and conditional aggregation structures within architectural frameworks. The experiments were conducted utilizing medium-sized models on images with a resolution of 256x256.}
\label{tab:ab3}
\vspace{-1.0em}
\end{table}

\begin{table}[h]
\centering
\renewcommand\arraystretch{1.2}
\begin{adjustbox}{width=0.5\textwidth,center}
\begin{tabular}{c|cccc} 
\toprule
Method & FID($\downarrow$) & LPIPS($\downarrow$) & SSIM($\uparrow$) & PSNR($\uparrow$) \\ 
\midrule
three conditional branch & 8.944 & 0.1688 & 0.7242 & 17.427 \\
Ours & 8.492 & 0.1687 & 0.7236 & 17.389 \\
\bottomrule
\end{tabular}
\end{adjustbox}
\caption{The study presents a comparative performance analysis of condition branches within the conditional aggregation component of the architecture. The experiments were conducted utilizing medium-sized models on images with a resolution of 256x256.}
\label{tab:ab4}
\vspace{-1.0em}
\end{table}

\subsection{Case Study}

\begin{figure}[h]
  \centering
   \includegraphics[width=0.75\linewidth]{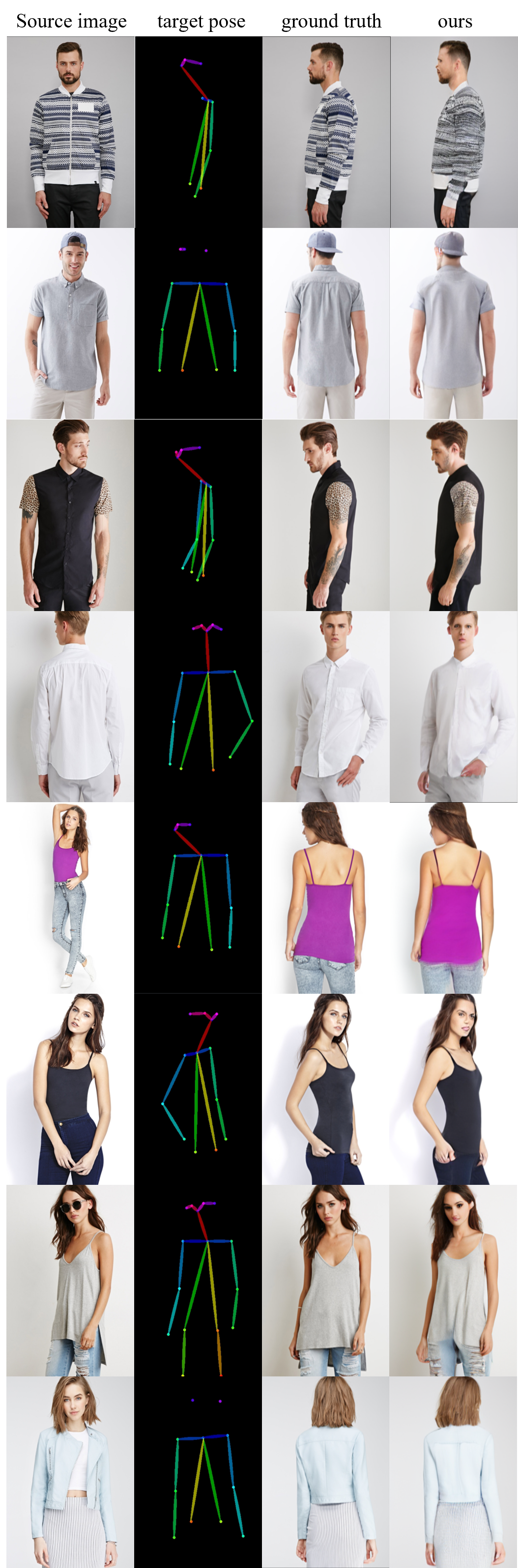}

   \caption{Our model produces images that correspond to the intended pose while maintaining the characteristics of the source image.}
   \label{fig:case}
\end{figure}

We conducted a comparative analysis of the images generated by the RPFM with the ground truth, as illustrated in \cref{fig:case}. The generated human images demonstrated high quality across various scenarios. By utilizing the flow matching model and transformer network, we efficiently sampled images while successfully adhering to the target pose requirements and partially preserving the appearance of the source image. This indicates that the alignment between the source image and the target pose was effectively achieved, resulting in the accurate production of the target image.

Upon a detailed examination of the figure, Row 1 reveals minor variations in the colors of clothing patterns; however, the generated image successfully aligns with the target pose while preserving certain visual attributes of the source image. In Row 2, the model adeptly inferred the back view of the clothing, which was not visible in the source image. Row 3 demonstrates that tattoos and other features were effectively preserved. Furthermore, in Rows 4, 5, 6, 7, and 8, both the target pose and the appearance of the source image were consistently maintained.

%% file: sec/5_conclusion.tex
\section{Conclusion}
\label{sec:conclusion}

\subsection{Conclusion}

In this paper, we introduce a novel methodology termed Real-time Person Image Synthesis using a Flow Matching Model (RPFM), designed to enhance the efficiency of Pose-Guided Person Image Synthesis and facilitate real-time video generation on a frame-by-frame basis. Our approach incorporates a flow matching model within the latent space and utilizes a transformer network, allowing for the adjustment of the NFE (Averaged Number of Function Evaluations) to expedite the sampling process. This enables the generation of images that retain the target pose while preserving significant visual attributes of the source image across numerous source image and target pose pairs. Through comprehensive experiments, we have demonstrated that our model achieves performance on par with state-of-the-art models while offering the fastest sampling speed. Furthermore, a case study was conducted to assess the quality of the generated images. Nonetheless, the challenge of fully preserving the fine details of the source image persists as an unresolved issue.

Given our primary objective of achieving high-speed performance, we anticipate that this research will establish a foundation for practical applications, such as real-time sign language video synthesis.

\subsection{Limitations}

One limitation of our study is the inability to conduct the ablation study up to 500 epochs due to resource constraints. Nevertheless, we consider it reasonable to evaluate the performance of various architectures after training for 100 epochs and to select the optimal architecture based on these results. Additionally, we did not explore the model's speed and performance trade-offs across a broader range of NFEs. Furthermore, a more in-depth analysis is required to understand why the medium-sized DiT-based model achieved a superior FID score compared to the large-sized DiT-based model. Moreover, while previous studies suggest that UNet~\cite{ronneberger2015u}-based diffusion models (such as PCDM) are expected to exhibit significantly slower sampling speeds than transformer-based diffusion models (such as X-MDPT), we were unable to conduct a precise experiment to verify this due to resource limitations.

%% file: sec/X_suppl.tex
\clearpage
\setcounter{page}{1}
\maketitlesupplementary

\noindent\textbf{Qualitative comparative study with baselines}

\begin{figure*}
  \centering
  \includegraphics[width=1\linewidth]{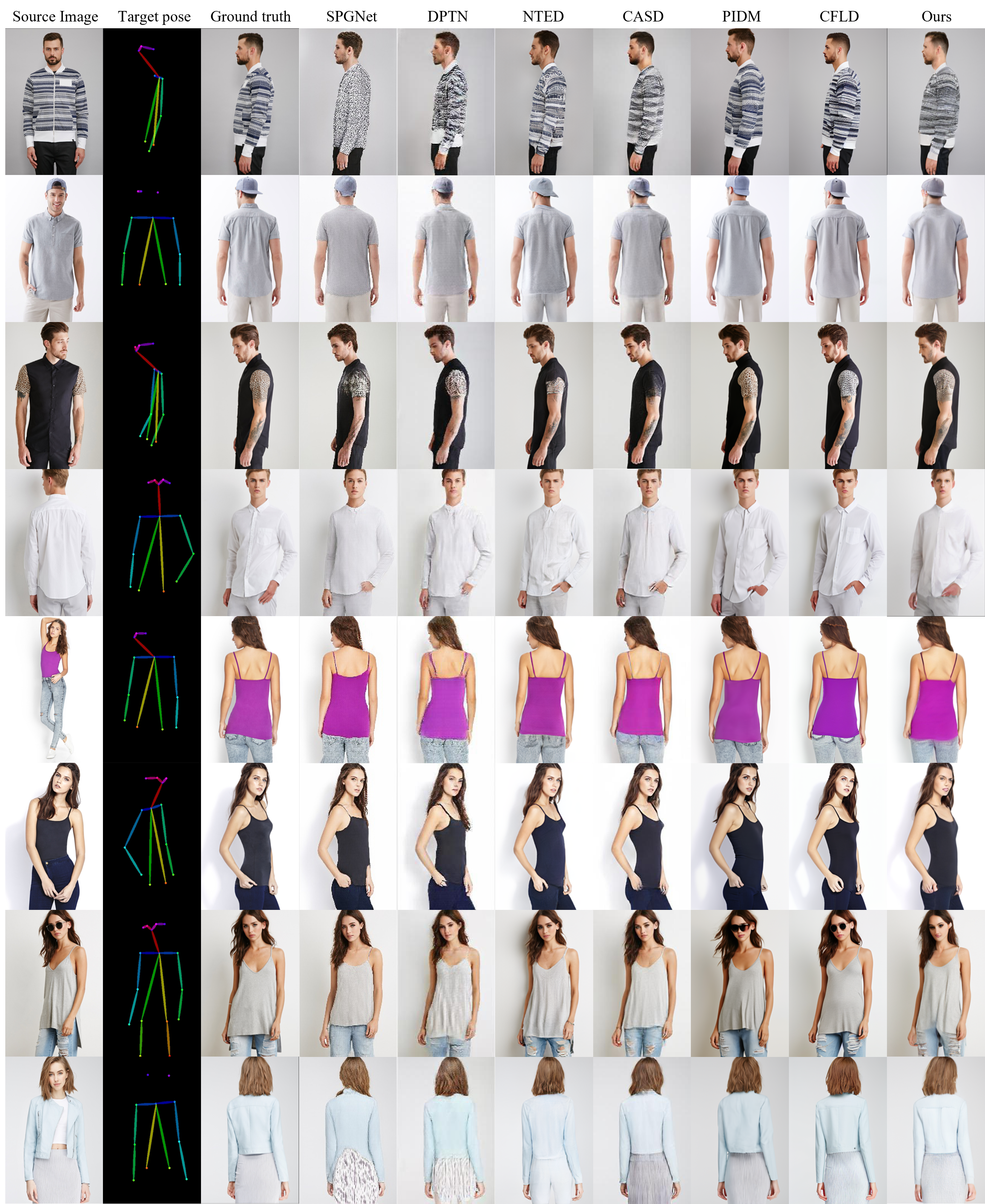}
  \caption{A qualitative comparative study with baselines (in the DeepFashion dataset)}
  \label{fig:suppl}
\end{figure*}

We present a qualitative comparative analysis utilizing baseline models, as illustrated in \cref{fig:suppl}. All experiments were conducted using the DeepFashion dataset at a resolution of 256x256. The generated images from baseline models, including SPGNet~\cite{lv2021learning}, DPTN~\cite{zhang2022exploring}, NTED~\cite{ren2022neural}, CASD~\cite{zhou2022cross}, PIDM~\cite{bhunia2023person}, and CFLD~\cite{lu2024coarse}, were sourced from the images available on the CFLD GitHub repository. Our model exhibits competitiveness with current state-of-the-art models and produces visually satisfactory generated images.